\documentclass[11pt]{article}

\usepackage[utf8]{inputenc} 
\usepackage[T1]{fontenc} 
\usepackage[numbers]{natbib}
\usepackage{geometry} 
\usepackage{graphicx}
\usepackage{subcaption}

\usepackage{graphicx} 
\usepackage{amsmath} 
\usepackage{booktabs} 
\usepackage{multirow} 
\usepackage{hyperref} 
\hypersetup{
    colorlinks=true,
    linkcolor=blue,
    filecolor=magenta,
    urlcolor=cyan,
}

\usepackage{titlesec}

\titleformat{\paragraph}
  {\normalfont\normalsize\bfseries}{\theparagraph}{1em}{}

\usepackage{listings} 
\usepackage{xcolor} 

\definecolor{codegreen}{rgb}{0,0.6,0}
\definecolor{codegray}{rgb}{0.5,0.5,0.5}
\definecolor{codepurple}{rgb}{0.58,0,0.82}
\definecolor{backcolour}{rgb}{0.95,0.95,0.92}

\lstdefinestyle{mystyle}{
    backgroundcolor=\color{backcolour},
    commentstyle=\color{codegreen},
    keywordstyle=\color{magenta},
    numberstyle=\tiny\color{codegray},
    stringstyle=\color{codepurple},
    basicstyle=\ttfamily\footnotesize,
    breakatwhitespace=false,
    breaklines=true,
    captionpos=b,
    keepspaces=true,
    numbers=left,
    numbersep=5pt,
    showspaces=false,
    showstringspaces=false,
    showtabs=false,
    tabsize=2
}
\hypersetup{
citecolor=black
}
\title{No Free Lunch in Language Model Bias Mitigation? Targeted Bias Reduction Can Exacerbate Unmitigated LLM Biases}
\author{
  Shireen Chand,$^*$ Faith Baca,$^*$ Emilio Ferrara  \\ 
  Thomas Lord Department of Computer Science, University of Southern California \\ 
  \texttt{shireenc@usc.edu; faithbac@usc.edu; emiliofe@usc.edu} 
}
\date{* equal contributors} 

\begin{document}

\maketitle 


\begin{abstract}
Large Language Models (LLMs) inherit societal biases from their training data, potentially leading to harmful or unfair outputs. While various techniques aim to mitigate these biases, their effects are often evaluated only along the dimension of the bias being targeted. This work investigates the cross-category consequences of targeted bias mitigation. We study four bias mitigation techniques applied across ten models from seven model families, and we explore racial, religious, profession- and gender-related biases. We measure the impact of debiasing on model coherence and stereotypical preference using the StereoSet benchmark. Our results consistently show that while targeted mitigation can sometimes reduce bias in the intended dimension, it frequently leads to unintended and often negative consequences in others, such as increasing model bias and decreasing general coherence. These findings underscore the critical need for robust, multi-dimensional evaluation tools when examining and developing bias mitigation strategies to avoid inadvertently shifting or worsening bias along untargeted axes.
\end{abstract}


\section{Introduction}
\label{sec:intro}

Large Language Models (LLMs) have become known as a widespread and revolutionary technology, embedded in many different applications that influence how we access information, create content and interact with the digital world. However, their increasing adoption is accompanied by a fundamental challenge: LLMs trained on large corpora of human-generated content inherit and frequently exacerbate deeply ingrained societal prejudices regarding race, gender, religion and other sensitive categories \cite{weidinger2022taxonomy,gallegos2024bias}. The risk of these models reinforcing harmful stereotypes is a critical barrier to their safe and fair adoption, making the development of effective bias mitigation techniques a central focus of AI research \cite{blodgett2020language,weidinger2021ethical,ferrara2023should}.

Numerous mitigation strategies have been proposed, ranging from data debiasing and constrained decoding to fine-tuning and parameter editing \citep{Bolukbasi2016,Bordia2019,Lauscher2021,lin2024towards}. However, the evaluation of these techniques often focuses narrowly on the specific bias dimension being targeted for reduction. Less understood are the potential side effects or cross-category impacts: for example, how does attempting to mitigate gender bias affect religious bias, or how does targeting race bias influence profession-related stereotypes?

This paper addresses this gap by systematically investigating the cross-category effects of several common mitigation strategies. Our research question is: \textbf{How does the mitigation of bias along a single axis (e.g., gender) affect the model's performance along several axes (gender, profession, religion, and race)?} Our work is motivated by this fundamental question, operating under the hypothesis that there is \textbf{No Free Lunch} in language model bias mitigation: We hypothesize that, due to the entangled nature of conceptual representation with LLMs, targeted interventions on singular bias dimensions will inevitably cause unintended side effects on other, unmitigated bias dimensions. To test this hypothesis, we propose and implement a \textbf{comprehensive auditing framework} for Transformer-based LLMs. 

\begin{figure}
    \centering
    \includegraphics[width=1\linewidth]{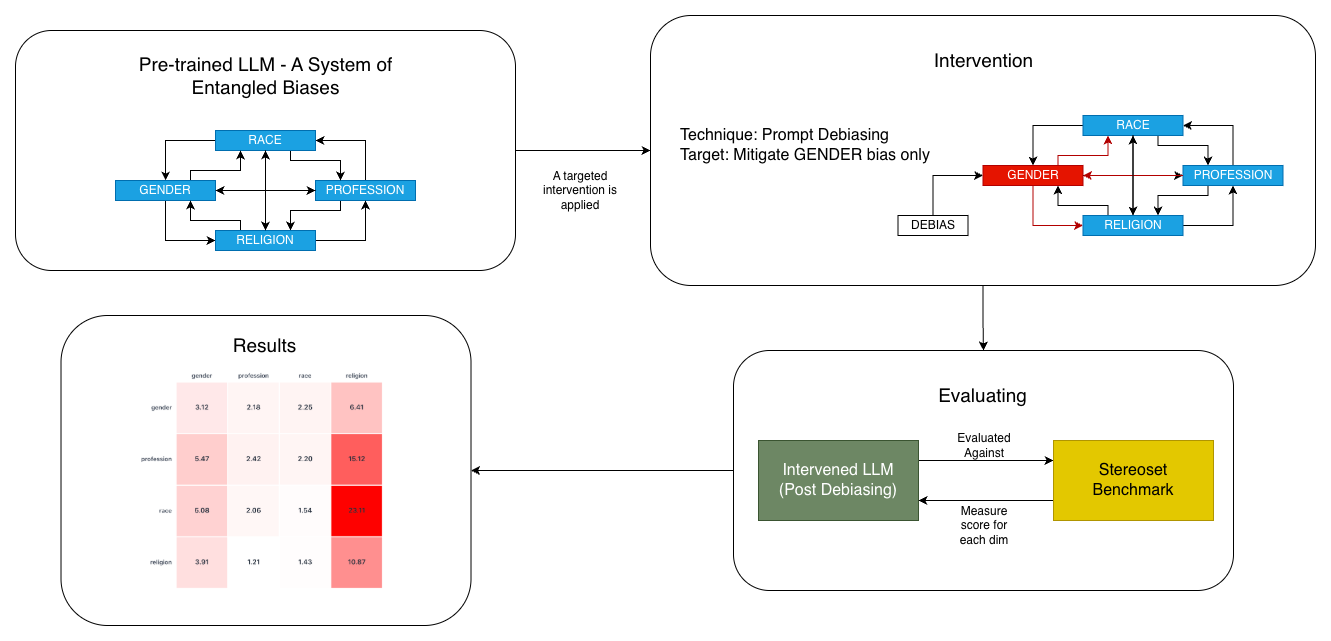}
    \caption{\textbf{A Visual Representation of Our Auditing Framework and the "No Free Lunch" Principle.} The process begins with a pre-trained LLM with entangled biases. A debiasing technique is applied to a single target dimension. The intervened model is then evaluated across all dimensions using the StereoSet benchmark.}
    \label{fig:framework}
\end{figure}

Our contributions are as follows:

\begin{itemize}
    \item We conduct a comprehensive study of four post-hoc debiasing techniques (Logit Steering, Activation Patching, BiasEdit and Prompt Debiasing) across ten language models, creating a robust and generalizable body of evidence.
    \item We find consistent and statistically significant evidence for our "No Free Lunch" hypothesis. Targeted debiasing frequently causes biases to spill over into untargeted dimensions, in some cases causing more harm than the original intervention sought to fix.
    \item We present our methodology as a necessary framework (cf., Figure \ref{fig:framework}) for the responsible evaluation of bias mitigation techniques, advocating for the adoption of multi-dimensional analysis as a standard practice in the field to prevent the inadvertent "trading" of one bias for another.
\end{itemize}

\section{Related Work}

\subsection{Trade-offs in Algorithmic Success and AI}
Efforts to improve fairness in LLMs often reveal just how intertwined linguistic structures and representations are within models. While the study of bias and fairness in machine learning is a well-established field, the notion of interconnectedness extends beyond the scope of LLMs and bias research specifically; adjustments to certain specific representational components in complex  technological systems inevitably lead to unforeseen trade-offs \cite{kleinberg2017inherent}. 

This challenge is fundamentally related to the \textbf{"No Free Lunch"} (NFL) theorem for optimization originally proposed by Wolpert and Macready in 1997 \cite{wolpert-nfl}. The NFL theory states that, for a given search or optimization algorithm, any gains in performance on one class of problems are necessarily offset by losses in performance on another class of problems. When applied to machine learning, the theory suggests that no single, universally superior algorithm or intervention exists for any type of problem, as improvements in one aspect of a system often come at the expense of another. 
We furthered this idea in the context of AI systems by positing the problem of "\textbf{Butterfly Effect}" in AI bias: small, targeted interventions can trigger cascading and unpredictable consequences in the broader system's behavior \cite{ferrara2024}. Though their original contexts are much wider in scope, both the Butterfly Effect and the NFL theory work in concert to offer a theoretical lens for the analysis of intervention trade-offs in bias mitigation.

\subsection{Bias Benchmarks}
Various recent surveys \cite{mehrabi2021,pessach2022,ferrara2024fairness} offer extensive overviews of the different sources of bias, starting from historical representation in training data to algorithmic processing and the different mathematical definitions of fairness. 

To quantify these biases, a variety of benchmarks have been developed. For example, datasets like CrowS-Pairs \cite{nangia2020} and WinoBias \cite{zhao2018} measure bias through paired sentences that differ only by a demographic term; the BOLD dataset \cite{dhamala2021} evaluates bias in open-ended text generation across a vast number of prompts. Bias evaluation has also been extended into other realms such as question answering \cite{parrish2022bbq} and Vision Language Models \cite{wang2024vlbiasbenchcomprehensivebenchmarkevaluating}. 

Our work adopts the StereoSet benchmark \cite{Nadeem2021}, which is uniquely suited to our research goals. Unlike binary choice datasets, StereoSet presents example contexts paired with triplets of sentences (stereotype, anti-stereotype, unrelated) which allows for the disentanglement of a model's linguistic coherence from its stereotypical preference. This is critical as Wang \textit{et al}. \cite{wangtradeoffs} find that there are significant trade-offs between fairness and accuracy in contexts like multi-task learning. Furthermore, it has been shown that catastrophic forgetting \cite{Kirkpatrick_2017} is a significant challenge for neural networks and LLMs in both learning and unlearning tasks \cite{nguyen,ouyang2022traininglanguagemodelsfollow,halevy-etal-2024-flex}; thus, measuring how debiasing affects model coherence is of essence. Additionally, StereoSet's multi-dimensional nature, covering race, gender, religion and profession, is also a prerequisite for our investigation into the cross-dimensional effects of bias mitigation.

\subsection{Existing Mitigation Techniques}
A significant body of work has focused on mitigating bias during the model's initial training or a subsequent full fine-tuning phase. These methods aim to embed fairness more fundamentally into the model's parameters. Techniques include data augmentation with counterfactual examples \cite{zmigrod2019}, re-weighting training examples to reduce the influence of biased data, and resource-intensive methods like Reinforcement Learning from Human Feedback (RLHF) to steer models toward less harmful behavior \cite{ouyang2022training,Bai2022}. Architectural analyses as done by Leteno \emph{et al.}  \cite{leteno2022}, try to identify the specific components responsible for encoding bias and give insights that can inform future model design. The Fair Class Balancing technique by Yan and collaborators \cite{yan2020} demonstrates a method for rebalancing training data not on the sensitive attributes themselves, but on automatically discovered proxy attributes, hence improving group fairness. While potentially more robust, these methods are highly computationally expensive, require access to large datasets, and involve a full training pipeline. 

A promising middle ground between full retraining and pure inference time methods is the field of model editing. These techniques make surgical, computationally efficient modification to the weights of a pre-trained model to alter a specific behavior. Techniques like ROME \cite{Meng2022}, MEMIT \cite{meng2023} and \textbf{BiasEdit} \cite{xu-etal-2025-biasedit} fall under this category. They are a better alternative to full fine-tuning, but still require direct access to the model's parameters. 

In contrast to training based methods, inference based (or post hoc) techniques are computationally cheap and model agnostic. The foundational idea of representing social bias as a linear direction in an embedding space was introduced by Bolukbasi \emph{et al.}  \cite{Bolukbasi2016} for static word embeddings. The authors demonstrated that certain biases, such as gender, could be identified via PCA on the difference vectors of definitional pairs (e.g., "he" vs "she") and subsequently removed via geometric projection. The \textbf{Logit Steering} and \textbf{Activation Patching} techniques are direct applications of this projection method to the hidden states of modern Transformer models. However, Hila Gonen and Yoav Goldberg \cite{gonen-goldberg-2019-lipstick} find that while these techniques may successfully remove the projection of bias, they tend to  leave the clustering of biased concepts intact in the vector space. Thus, debiasing may often operate at a superficial level and, while effective on the surface, may fail to eliminate the underlying structures responsible for bias emergence. 

While the aforementioned techniques have been shown to be generally effective at reducing bias on their target dimension, their evaluation on untargeted dimensions is often overlooked. Most studies measure the reduction of a specific bias and may track its effect on overall model capabilities like perplexity. However, the potential for collateral damage where an intervention on one bias axis inadvertently introduces or exacerbates bias on another is a critical but underexplored area. This paper directly addresses this gap. By applying a suite of interventions and measuring their effects across all four StereoSet dimensions, we provide a complete analysis of their true costs, suggesting that the "\textbf{No Free Lunch}" principle is at play.

\section{Methodology}
\label{sec:methodology}

\subsection{Dataset: StereoSet}
To evaluate model bias across multiple domains, we utilize the \textbf{StereoSet} datatset \cite{Nadeem2021}, a widely recognized benchmark designed specifically to measure stereotypical biases in pretrained language models. The dataset's structure is uniquely suited for our research, as it allows for a nuanced quantification of bias rather than just its detection. 

For our experiments, we use the \textbf{intersentence} portion of StereoSet data and the benchmark's accompanying metrics to probe models' coherence and preferences for stereotypical sentences.

\subsubsection{Scope and Dimensions}
The StereoSet dataset is organized into four distinct domains of social bias, allowing us to investigate the central thesis of our paper: that mitigating bias in one dimension can exacerbate it in others. The validations set used in our evaluation contains a total of \textbf{2123} examples, distributed across four dimensions (race, gender, religion, profession). More specifically, after we apply very minor filtering, there are 242, 827, 976, and 78 gender, profession, race, and religion intersentence entries, respectively.

\subsubsection{Structure of an Intersentence StereoSet Entry}

Each data point consists of a \textbf{context sentence} and a set of three carefully constructed \textbf{stereotype, anti-stereotype, and unrelated completions}. The stereotype completions are designed to reflect existing social stereotypes while the anti-stereotype completions subvert these biases. The unrelated completions are included to evaluate the linguistic coherence of models. 

An example entry taken from the StereoSet data's race category is as follows:
\begin{itemize}
    \item \textbf{Context}: "My neighbor is Hispanic."
    \item \textbf{Stereotype}: "He doesn't speak English."
    \item \textbf{Anti-stereotype}: "He went to college and is a teacher." 
    \item \textbf{Unrelated}: "Dogs have funny tails."
\end{itemize}

\subsubsection{Evaluation Metrics}
The \textit{Language Modeling Score} (LMS) serves as a measure of a model's fundamental linguistic competence. In the context of intersentence StereoSet data, LMS evaluates a model's ability to distinguish between meaningful and unrelated sentence completions. For each example, a model is considered successful if it assigns a higher probability to either the stereotypical or anti-stereotypical completion than to the unrelated completion. The final score is the percentage of examples where this condition is met. 

The LMS is calculated as:

\[
\text{LMS} = 100 \times \frac{\sum_{i=1}^N I (max(P_{stereo,i}, P_{anti-stereo,i}) > P_{unrelated,i})} {N}
\]

where \(N\) is the total number of examples and \(I\) is the indicator function. For each example \(i\), \(P_{\text{stereo},i}\), \(P_{\text{anti-stereo},i}\), and \(P_{\text{unrelated},i}\) denote the model-assigned probabilities of the stereotypical, anti-stereotypical, and unrelated completions, respectively. A higher LMS indicates a more coherent model that better understands the context, and the LMS of an ideal model is 100.

The next evaluation metric is the \textit{Stereotype Score} (SS), which directly measures the model's bias by quantifying its preference for stereotypical associations. It is calculated as the percentage of examples in which the model assigns a higher probability to the stereotypical completion over the anti-stereotypical one.

The SS is calculated as:

\[
\text{SS} = 100 \times \frac{\sum_{i=1}^N I(P_{stereo,i} > P_{anti-stereo,i})}{N}
\]

A score of 100 indicates a complete preference for stereotypical assocations, while a score of 0 indicates a complete preference for anti-stereotypical ones. An ideally unbiased model would demonstrate no preference, yielding an SS of 50.

\subsubsection{Idealized CAT Score (ICAT)}
To provide a single, holistic measure that balances linguistic competence with fairness, we use the \textit{Idealized Correlation Association Test} (ICAT) score. The ICAT score combines  LMS and SS, rewarding models that are both knowledgeable (high LMS) and unbiased (SS close to 50). 

The score is formulated to penalize models that are biased in either the stereotyical or anti-stereotypical direction through its fairness component, \(\frac {min(SS, 100-SS)}{50} \). This term is maximized at 1 when SS is 50 and drops to 0 when SS is either 0 or 100.

The ICAT score is calculated as:

\[
\text{ICAT} = LMS \times \frac {min(SS, 100-SS)}{50}
\]

The ICAT score ranges from 0 to 100 and satisfies several desirable axioms:

\begin{itemize}
    \item An ideal model with perfect coherence (LMS=100) and no bias (SS=50) achieves an ICAT score of 100.
    \item A fully biased model (SS=0 or SS=100) achieves an ICAT score of 0, regardless of its LMS.
    \item A random-guess model (LMS=50, SS=50) achieves an ICAT score of 50.
\end{itemize}

\subsection{Models}
In order to comprehensively evaluate bias and mitigation techniques, we conduct experiments across a diverse set of transformer-based LLMs. This range allows us to observe how model features influence baseline biases and to quantify the efficacy of bias mitigation methods. 
The models used in our study are summarized in Table~\ref{tab:models}.

\begin{table}
\centering
\begin{tabular}{l l l}
\hline
    \textbf{Family} & \textbf{Model} & \textbf{Parameters} \\
    \hline
    Gemma & \textit{google/gemma-2b} & 2B \\
    Gemma & \textit{google/gemma-7b} & 7B \\
    OLMo & \textit{allenai/OLMo-1B-0724-hf} & 1B \\
    OLMo & \textit{allenai/OLMo-2-1124-7B} & 7B \\
    LLaMA & \textit{meta-llama/Llama-3.2-1B} & 1B \\
    LLaMA & \textit{meta-llama/Llama-2-7b-hf} & 7B \\
    Qwen & \textit{Qwen/Qwen2.5-3B-Instruct} & 3B \\
    GPT-Neo & \textit{EleutherAI/gpt-neo-1.3B} & 1.3B \\
    Mistral & \textit{mistralai/Mistral-7B-Instruct-v0.3} & 7B \\
    Deepseek & \textit{deepseek-ai/deepseek-llm-7b-chat} & 7B \\
\hline
\end{tabular}
\caption{Models Used in Experiments}
\label{tab:models}
\end{table}

These models were chosen to represent a broad spectrum of model characteristics including size and architecture. The selection enables us to determine whether bias mitigation techniques are more effective for certain models, and to analyze how cross-category bias spillover manifests across different LLM families. 

\subsection{Bias Mitigation Techniques}
To investigate the trade-offs of targeted debiasing, we implement four distinct techniques. While all are applied post-hoc without full retraining, they represent three different families of intervention: \textbf{Geometric Interventions} that manipulate activations in-flight, \textbf{Model Editing Interventions} that make surgical modifications to model weights, and \textbf{Input-Based Interventions} that modify the prompt.

\subsubsection{Bias Direction Computation via PCA} \label{sec:pca}

To perform targeted interventions, we must first represent an abstract bias concept as a concrete direction in the model's activation space. We adopt the methodology pioneered by Bolukbasi and collaborators \cite{Bolukbasi2016} for word embeddings and adapt it for contextual language models.

We begin by selecting contrastive pairs for each bias dimension that represent the poles of the bias axis (e.g., ("He is", "She is") for gender, ("Black person", "White person") for race). Each text in a pair is fed through the model, and we extract the final-layer hidden state representations. The hidden state for each text is averaged across all token positions to produce a single vector.

For each pair, we compute the difference between the two resulting vectors (e.g., \(\vec{h}_{\text{"He is"}} - \vec{h}_{\text{"She is"}}\). This creates a set of difference vectors, each pointing along a slightly different instantiation of the bias axis. To find the single, most dominant direction of variance across all difference vectors, we perform \textbf{Principal Component Analysis (PCA)} and extract the first principal component.

The resulting vector is normalized to have a unit length, giving a pure directional vector, \(\vec{v}_{bias}\) that represents the \textbf{core axis of the targeted bias} within the model's activation space. The computed bias vector \(\vec{v}_{bias}\) serves as the basis for Logit Steering and Activation Patching.

\subsubsection{Geometric Interventions}

The following bias mitigation techniques operate by geometrically projecting out the pre-computed bias direction from the model's hidden state activations during the forward pass.

\paragraph{Logit Steering (Projection-Based Debiasing)}

Logit Steering is an inference-time intervention that aims to remove the influence of the bias direction from the model's activations at a specific layer. The technique is implemented via a forward hook attached to the penultimate layer of the model. 

During the forward pass, for each hidden state vector \(\vec{h}\) produced by this layer, we perform a linear projection to remove the component that aligns with the bias direction:

\[
\vec{h}_{debiased} = \vec{h} - \alpha \cdot \text{proj}_{\vec{v}_{bias}}(\vec{h})
\]

where \(\text{proj}_{\vec{v}_{bias}}(\vec{h})\) is the projection of \(\vec{h}\) onto the bias vector \(\vec{v}_{bias}\). The hyperparameter \(\alpha\) is a scaling factor that controls the strength of the intervention. For all our experiments, we use \(\alpha = 1.0\), representing a full removal of the projected bias component.

\paragraph{Activation Patching}

Activation Patching is a more aggressive form of projection-based debiasing, repurposing a technique from the field of mechanistic interpretability \cite{Meng2022}. While the mathematical operation is identical to Logit Steering, the key difference is its scope \cite{zhang2023towards}. 

Based on the hypothesis that bias is a distributed representation across multiple layers, this technique applies the same "projection and subtraction" intervention simultaneously to several layers of the model. We apply the hook to the final \textbf{five} layers of each model. This constitutes a more forceful patch intended to remove the bias signal more thoroughly as it propogates through the network. As with Logit steering, we use an intervention strength of \(\alpha = 1.0\).

\subsubsection{Prompt Debiasing (Input based Intervention)}

Unlike the geometric interventions, Prompt Debiasing is an input-based technique that uses the instruction-following capabilities of modern LLMs \cite{schick2021self,Bai2022}. This method involves prepending a carefully crafted instructional prompt to the original StereoSet context before it is processed by the model.

The prompts explicitly instruct the model to avoid the targeted bias and provide examples of neutral behavior. For example, the prompt for the gender dimension begins: 

\textit{"Please respond without gender stereotypes or bias. Treat all genders equally. Examples of unbiased statement: - An engineer could be any gender...."}

A critical aspect of our methodology is the evaluation of this technique. To ensure we are only measuring the model's performance on the actual task completion, the tokens corresponding to our instructional prompt are masked and excluded from the Negative Log-Likelihood (NLL) calculation. This isolates the effect of the conditioning prompt on the model's output quality.

\subsubsection{BiasEdit (Parameter Editing)}
BiasEdit is a targeted model editing approach that modifies a small subset of a model’s parameters to reduce stereotypical bias while preserving overall language modeling performance \cite{xu-etal-2025-biasedit}. The method has been shown to successfully reduce racial, religious, and gender-related biases in transformer-based LLMs while minimally affecting downstream task performance, but there is no investigation of its cross-dimension effects. 

The technique works by employing \textbf{lightweight editor networks} that generate parameter updates for specific model components. Based on preliminary bias tracing experiments, Xu \emph{et al.}  \cite{xu-etal-2025-biasedit} conclude that stereotypical associations tend to be concentrated in the MLP layers of transformer blocks with co-occurrences being captured in lower layers. Additional results determine which specific layers are optimal for debiasing. For consistency, we implement the technique on the \textbf{penultimate} layer of each model to balance intervention effectiveness with minimal disruption to overall model performance. 

To train the editor networks, we utilize the same StereoSet examples as in other methods with a \textbf{8:1} train-dev split for each dimension. The editing process is guided by two loss functions: a symmetric debiasing loss that encourages models to assign equal probability to StereoSet's stereotypical and anti-stereotypical completions, and a retention loss that preserves language modeling capabilities by attempting to maintain predictions on neutral completions. Thus, critically, BiasEdit's goal is  not simply to reduce stereotypical bias within models, but to achieve equal distributions between stereotypical and anti-stereotypical predictions while maintaining coherence. While the approach is defined for intrasentence data, we adapt it to handle StereoSet's intersentence examples to reflect our goal of evaluating and understanding the manifestation bias across complex contexts. 
This process yields a model specifically adapted based on anti-stereotypes from a single bias dimension.


\section{Auditing Framework}
\label{sec:procedure}

\subsection{Stage 1: Baseline Performance Calculation}

The initial and most important phase is the establishment of a performance baseline for each model. This provides the reference point against which all changes are measured. The pre-trained language model is loaded and run on the StereoSet dataset without any debiasing interventions active. The evaluation is performed independently for each of the four bias dimensions. The raw LMS, SS and ICAT scores for each dimension are calculated and saved.

\subsection{Stage 2: Intervention Application and Evaluation}
For the geometric techniques, the bias direction vector \((\vec{v}_{bias}\)) for the target dimension is computed using PCA as described in section \S~\ref{sec:pca}. For BiasEdit, the necessary weight modifications are calculated. Next, the specific debiasing technique is activated. For Logit Steering and Activation Patching, the appropriate forward hooks are registered on the model's layers. For BiasEdit, the pre-calculated weight changes are applied to the model. For Prompt Debiasing, the relevant instructional prompt is prepared for prepending to the input.

\subsection{Stage 3: Multi-Dimensional Evaluation}
With the intervention active for the chosen target dimension, the model is evaluated on the StereoSet benchmark across all four evaluation dimensions using LMS, SS, and ICAT. This process is designed to capture not only the intended effects but also the unintended collateral damage central to our "No Free Lunch" thesis.


\section{Results}
\label{sec:results}

We conducted \textbf{160 unique debiasing experiments}, evaluating the language models across 4 techniques and 4 target dimensions. Each experiment was audited by measuring its impact across all 4 fairness dimensions, resulting in \textbf{640 total evaluations}. Using ICAT score as our measure of a model's overall utility, our results reveal that targeted interventions achieved a statistically significant improvement in the on-target ICAT score in only \textbf{20.6\%} cases. Conversely, these same interventions caused statistically significant collateral damage, worsening the ICAT score on unmitigated, spillover dimensions in \textbf{31.5\%} of all spillover evaluations.

\subsection{The "No Free Lunch" Principle: A Systemic View of Trade-offs}

\begin{figure}[t]
    \centering
    \includegraphics[width=1\linewidth]{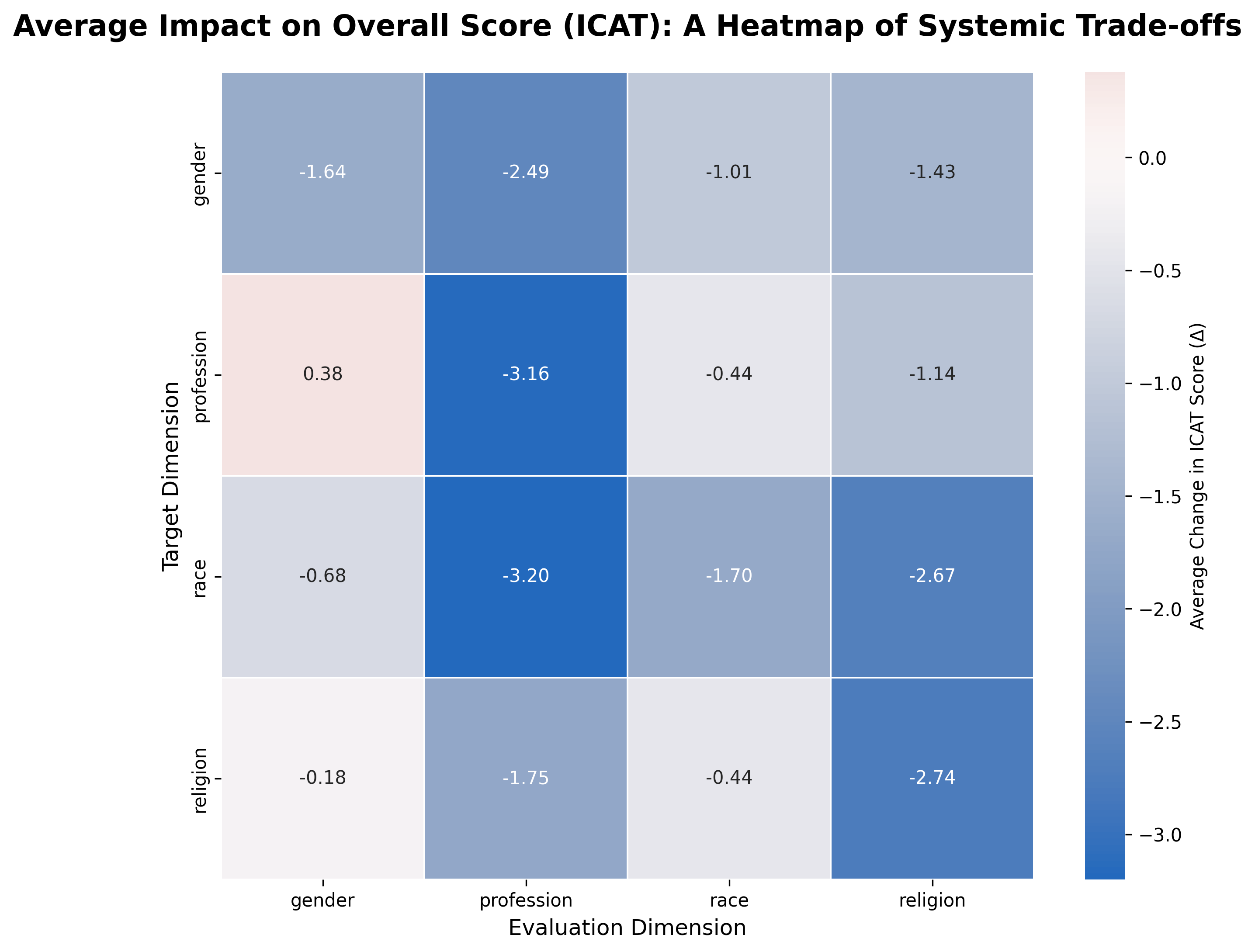}
    \caption{\textbf{Average Impact on Overall Score (ICAT)}. Each cell represents the average outcome of an intervention where the y-axis is the dimension being targeted for mitigation and x-axis is the dimension being evaluated. Blue cells indicate a negative average change (net harm to the model's quality and fairness), while red cells indicate a postive change (net improvement).}
    \label{fig:avg_icat_heatmap}
\end{figure}

Our primary finding is that bias mitigation is not a localized fix but a systemic intervention with far-reaching consequences. The Heatmap in Figure \ref{fig:avg_icat_heatmap} summarizes this phenomenon by showing the average change in the model's overall utility (ICAT score) for every target-evaluation pair.

The results are striking and reveal the potential for a pattern of \textbf{systemic harm}. The most dominant feature is the prevalence of negative (blue) values, indicating that these interventions, on average, damage the model's overall quality. This is true not only for off-target "collateral damage" but for the on-target intervention itself. 

For example, consider the case where \textit{profession} is both the target and evaluation dimension, which shows a catastrophic average ICAT drop of \textbf{-3.16}, a result that is statistically significant (t(39) = -2.22, p < .05). This means that the techniques applied to "fix" profession bias were so harmful to the model's core linguistic capabilities that they made the model significantly worse at handling the topic of professions: In essence, \textit{the cure was worse than the disease}. Similarly, targeting race bias led to an average on-target ICAT drop of \textbf{-1.70} while also causing significant collateral damage to the model's performance on profession with a drop of \textbf{-3.20} (t(39) = -2.28, p < .05). 

\begin{figure}[t]
    \centering
    \includegraphics[width=1\linewidth]{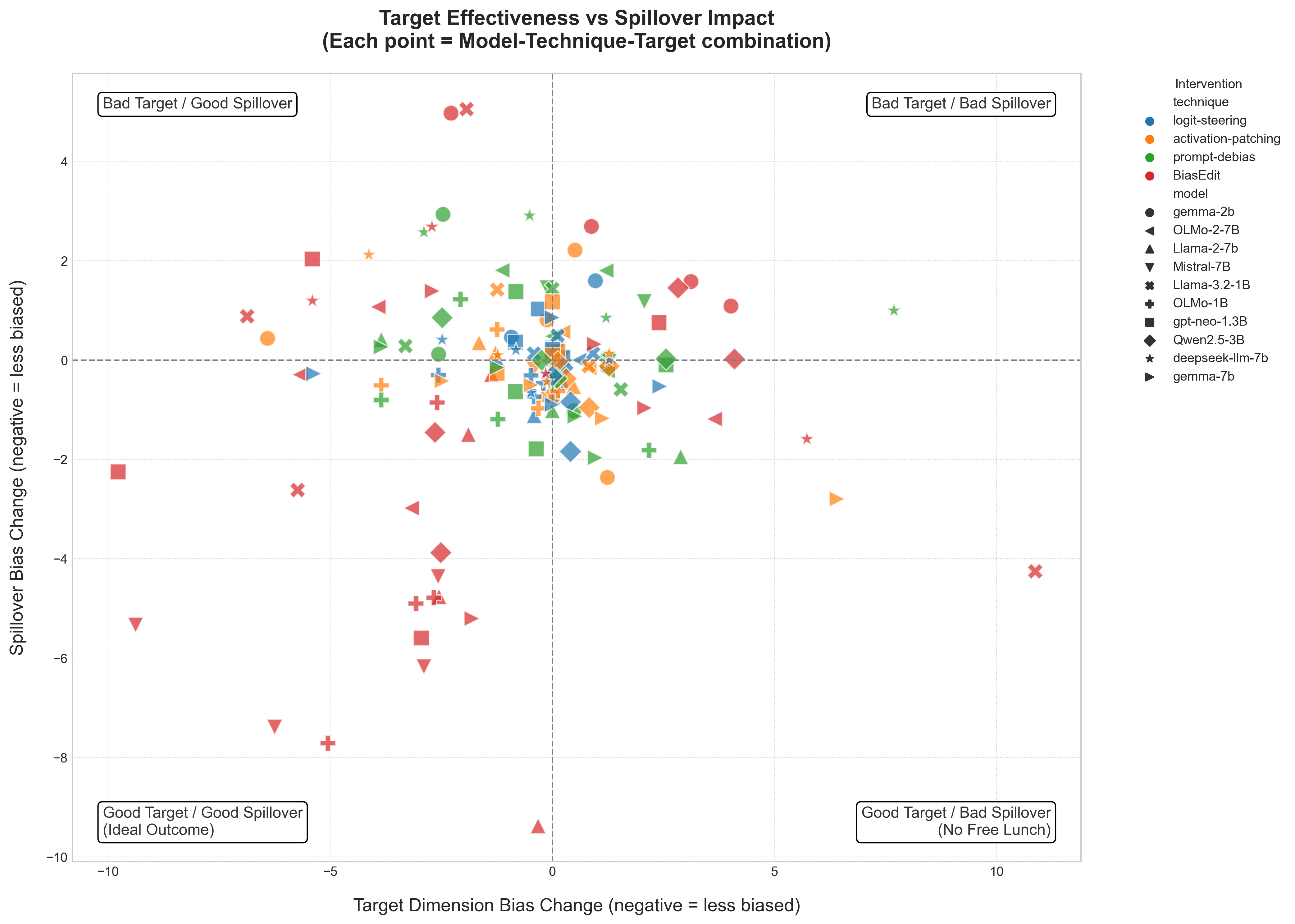}
    \caption{\textbf{Target Effectiveness vs Spillover Impact (Stereotype Change).} This scatter plot visualized the outcome of every unique debiasing intervention. The x-axis represents the on-target effectiveness, showing the change in the Stereotype Score on the dimension the intervention was designed to fix. The y-axis represents the collateral impact.}
    \label{fig:impact}
\end{figure}

The Scatter Plot in Figure \ref{fig:impact} confirms this is not an artifact of averaging. While some interventions land in the ideal outcome quadrant (bottom-left, i.e., \textit{Good Target / Good Spillover}), a dense cluster populates the \textbf{"No Free Lunch"} quadrant (bottom-right, i.e., \textit{Good Target / Bad Spillover}), showing that it is a frequent outcome for a successful stereotype reduction to be paid for with an increase in stereotypical associations elsewhere.

\subsection{Dimension-Specific Debiasing Success}
\begin{figure}[t]
    \centering
    \begin{subfigure}[b]{0.48\linewidth}
        \centering
        \includegraphics[width=1.123\linewidth]{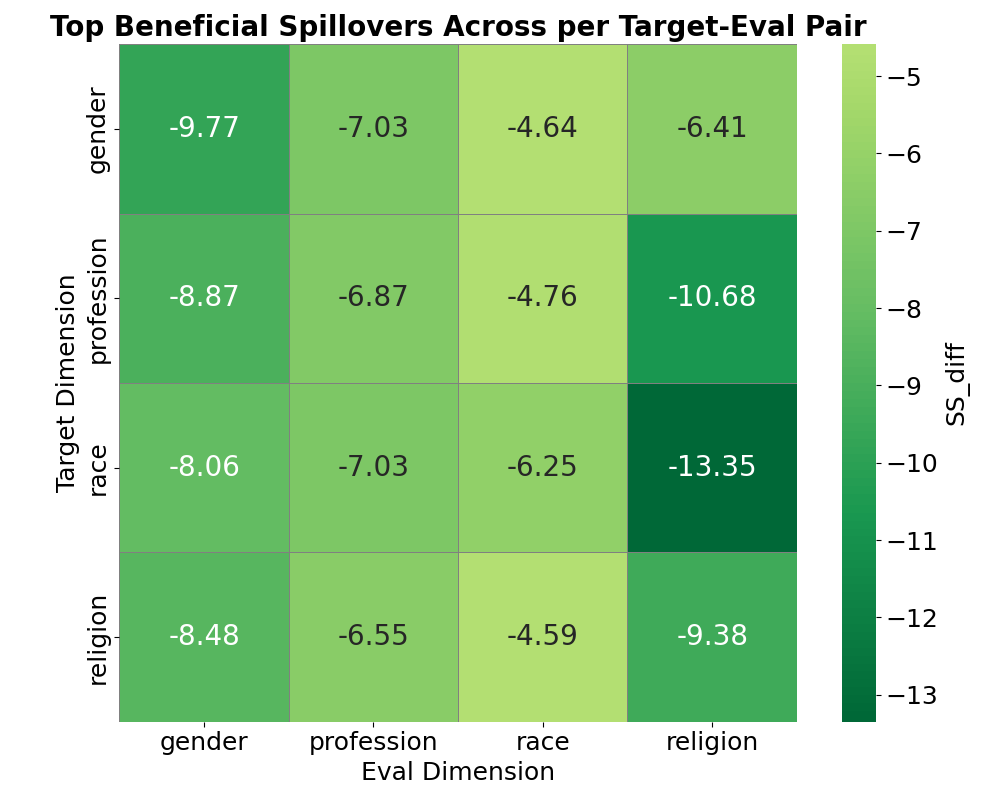}
        \caption{Beneficial Spillovers}
        \label{fig:positive_spillover}
    \end{subfigure}
    \hfill
    \begin{subfigure}[b]{0.48\linewidth}
        \centering
        \includegraphics[width=1.123\linewidth]{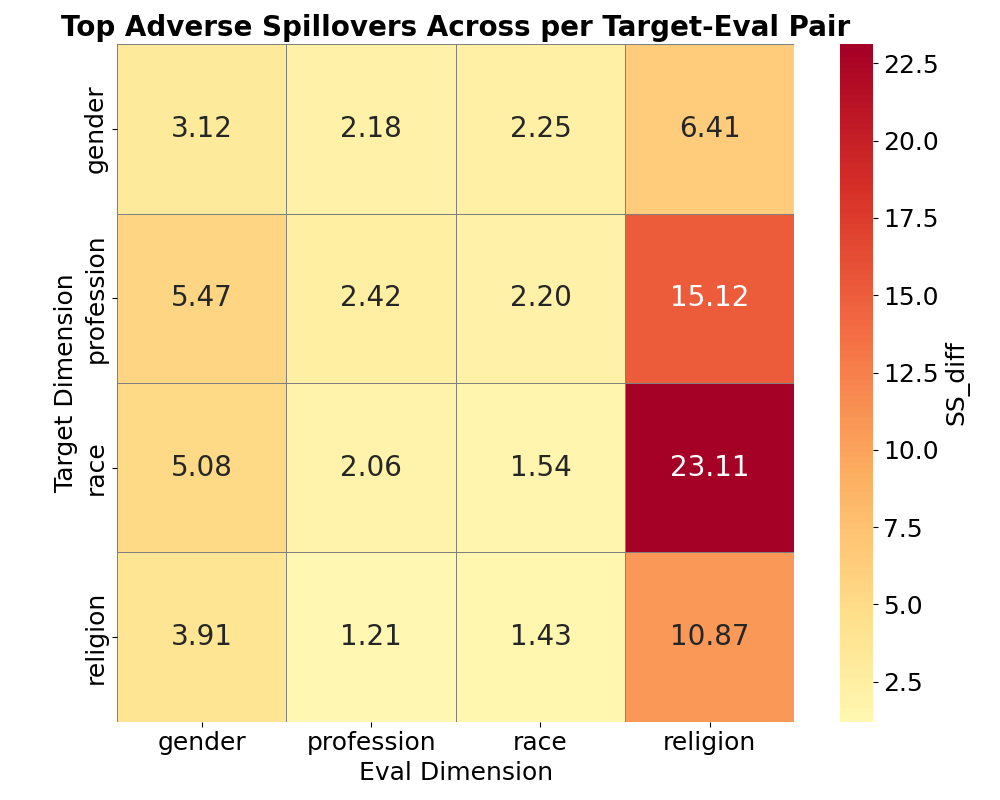}
        \caption{Adverse Spillovers}
        \label{fig:negative_spillover}
    \end{subfigure}
    \caption{Dimension-specific debiasing spillover effects, showing cases with beneficial and adverse spillovers (reductions and increases in LMS, respectively). Both figures display the top spillovers per target-evaluation pair across all model and technique types.}
    \label{fig:spillovers}
\end{figure}

Our analysis displays substantial variation in debiasing success across the four dimensions: some dimensions proved quite amenable to intervention while others resulted in significant increases in bias levels, as displayed in Figure \ref{fig:spillovers}. 

Religion emerged as the most spillover-susceptible evaluation dimension, exhibiting both the top beneficial and top adverse spillovers. This suggests that religion is highly entangled with other dimensions of bias and that models may lack the capability to representationally distinguish racial bias, for example, from religious bias. 

Gender as an evaluation dimension follows closely behind in terms of this pattern. Another potential explanation is that, since there are many fewer StereoSet triplets for both religion and gender, metrics become more sensitive to small changes thus amplifying observed spillovers. This highlights the importance of balance in dimensional composition in future efforts to create bias benchmarks. 

The beneficial spillovers warrant extra scrutiny as successful reduction of SS does not require that a model maintains its coherence. The top three most beneficial cross-category spillovers were the result of applying the BiasEdit technique. In two of these runs, applying the technique increased LMS, but in the last one, \textbf{LMS decreased by more than 20\%}. This shows that mitigating bias along one dimension can result in significant, unintended consequences unrelated to the main goal of debiasing --both good and bad-- along other dimensions.

The asymmetric pattern of spillovers suggests that bias mitigation techniques seeking to reduce bias along one dimension at a time may be insufficient. Real-world biases are complex and often represented intersectionally \cite{crenshaw1991mapping,weiintersectional} in LLMs \cite{ma-etal-2023-intersectional,souani2025hinterexposinghiddenintersectional}. Beyond the context of LLMs, Kearns \textit{et al}. \cite{kearns2018preventingfairnessgerrymanderingauditing} show that satisfying fairness constraints for certain independent groups does not guarantee fairness for their intersections. Thus, future work in fairness must address these concerns to accurately represent real-world biases. A potential avenue for exploration to reduce spillover effects is debiasing models sequentially so that dimensions are addressed in order of their independence of other dimensions. Addressing the issue of cross-dimension spillover is critical to ensure progress toward fairer LLMs.

\begin{figure}[h]
    \centering
    \includegraphics[width=1\linewidth]{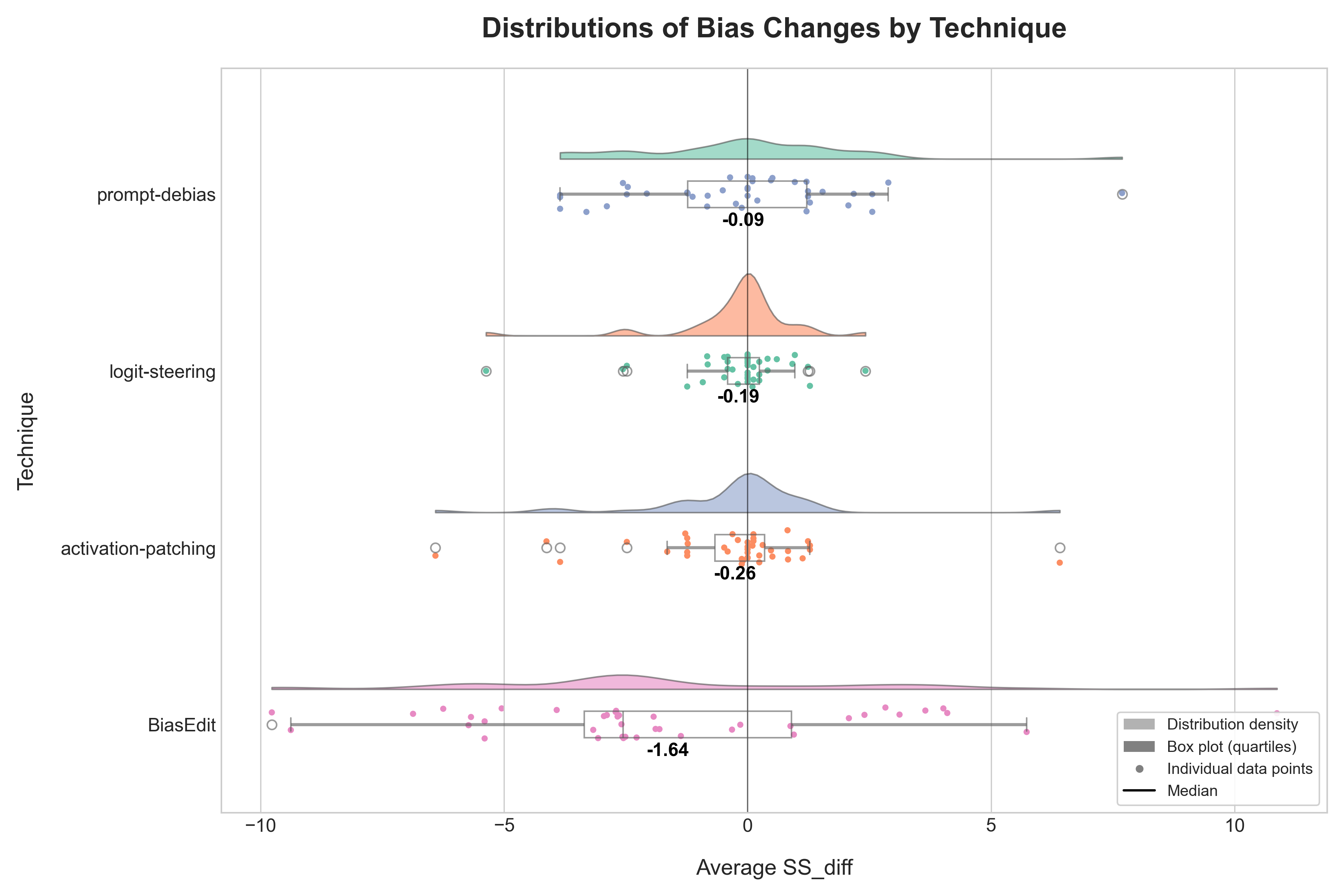}
    \caption{Change in bias is quantified through SS\_diff. SS\_diff changes averaged across all models are displayed. Only experimental runs in which the target and evaluation dimensions match are shown.}
    \label{fig:bias_change}
\end{figure}

\subsection{Analysis of Bias Mitigation by Technique}
In addition to examining the cross-dimension spillover effects, we analyze each technique's success in mitigating bias along intended dimensions. A successful experimental run is defined here as a reduction in SS. Additionally, Figure \ref{fig:bias_change} displays the distributions of change in SS across models for target dimension reduction. 

BiasEdit was overall the most successful debiasing technique, reducing SS along intended dimensions in 72.5\% of experimental runs. Nonetheless, Figure \ref{fig:bias_change} shows that BiasEdit also displayed the largest range in SS change by far, suggesting that, while the technique may be successful in reducing bias in many cases, its efficacy is highly model- and dimension-dependent. This contrasts the original results presented in \cite{xu-etal-2025-biasedit} of implementing the method on intrasentence data, indicating that intersentence complexity is also a significant factor in the technique's variability. Therefore, it is critical to develop debiasing methods that support intersentence data to reflect real-world language and biases more faithfully.

Logit-steering was the least successful method overall, reducing SS in only 35\% of runs. This suggests its intervention is often too weak to overcome the model's pre-existing biases. Activation Patching and Prompt Debiasing occupy a middle ground, succeeding 42.5\% and 45.0\% of the time, respectively, both contributing to modest average decreases in SS.

\subsection{Model Analysis}

\begin{figure}[h]
    \centering
    \includegraphics[width=1\linewidth]{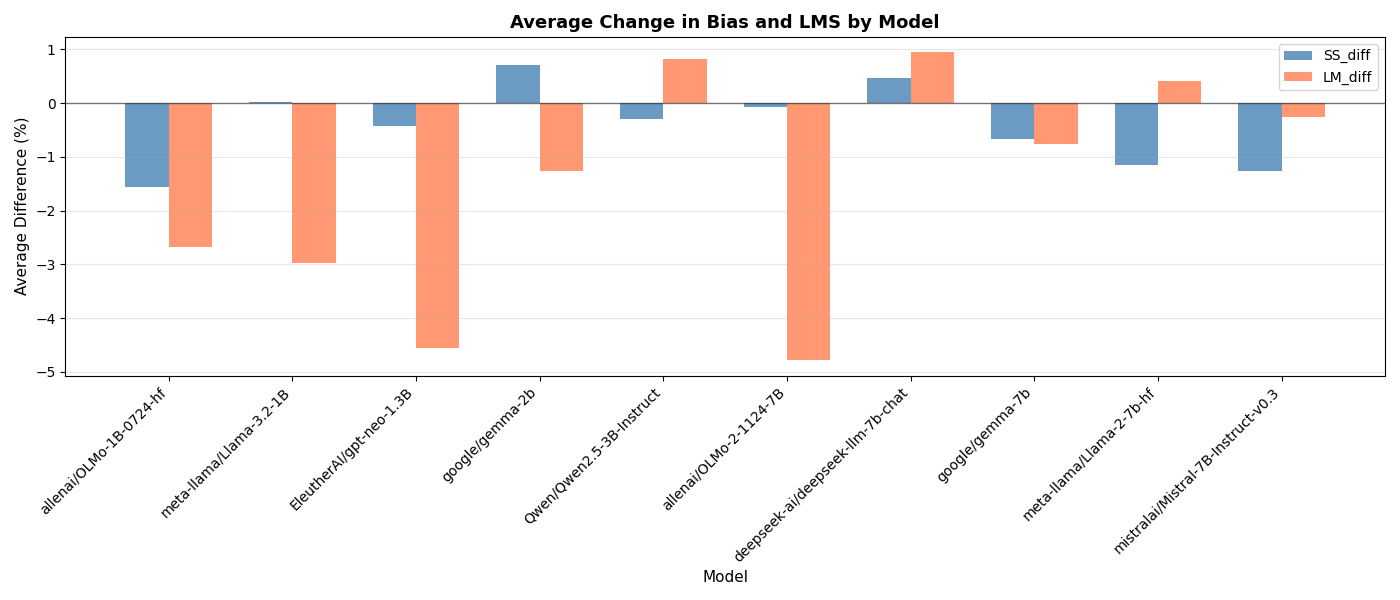}
    \caption{Change in SS and LMS metrics by model type. Averages are computed over all technique and dimension types.}
    \label{fig:models}
\end{figure}

To tie our analyses together, we examine how debiasing interacts with model architecture. Figure \ref{fig:models} displays the changes in SS and LMS averaged across technique and dimension types. The figure also makes clear that editing resulted changes in bias and model coherence that varied substantially between models. 

Generally, models with fewer parameters display larger drops in LMS, indicating that smaller models were much more susceptible to losses in coherence resulting from intervention. This is likely because smaller models rely more heavily on compact, intertwined representations of language, meaning that any slight perturbation --including debiasing along a singular dimension-- can be highly damaging to the internal structure responsible for general language modeling capabilities. Decreases in LMS occurred in seven out of the ten models after debiasing was applied. 

Both Gemma-2b and DeepSeek-7b displayed increases in SS after debiasing. These models may encode biases in ways that are inaccessible to our debiasing techniques. Qwen-3B presents a puzzling case: while LMS is decreased after intervention, SS is increased, implying that the model became both more biased and less coherent overall. This behavior shines light on the scarcity of our understanding about internal representations of bias in LLMs, and further work is needed to thoroughly assess how complex biases manifest in varying model architectures.

\section{Discussion}
\label{sec:discussion}

Our central finding that targeted interventions frequently cause harm to unmitigated dimensions can be explained by the \textbf{entangled nature} of conceptual representations within LLMs \cite{caliskan2017semantics}. Our results strongly suggest that a model does not learn "gender", "race", etc., as discrete, orthogonal concepts. Instead, these are overlapping, co-dependent subspaces learned from a training corpus where they are deeply linked. 

The vulnerability of the religion dimension to spillover is a prime example. In Western-based training data, discussions of religion are intertwined with gender roles, ethnic identities and specific professions \cite{kirk2021bias}. Consequently, when an intervention forcefully alters the model's representation of gender or race, it is not adjusting an isolated variable but disturbing a thread that runs through many other concepts. The resulting collateral damage is not a bug but an emergent feature of the entangled knowledge.  

Our results prove that evaluating a debiasing technique solely on its intended target is insufficient and misleading since it might appear successful but may be silently amplifying other harms. New auditing techniques for LLMs are emerging \cite{amirizaniani2024auditllm,QIU2026122702}: We argue that the use of a \textbf{multi-dimensional auditing framework} such as the one we proposed in this paper should become a standard practice. Before deploying any bias mitigation technique, practitioners must perform a comprehensive evaluation to map its full impact, measuring not only the intended effects but also the unintended spillover. 

Finally, we must acknowledge the limitations of our study. Our analysis is based on the StereoSet benchmark which, as pointed out by Blodgett and collaborators \cite{blodgett2021}, has some known limitations: They argue that fairness benchmarks like StereoSet inevitably encode a specific set of societal norms and stereotypes reflective of their place and time of creation, which in this case is modern, English-speaking cultures. The associations it labels as "stereotypical" may not be universally applicable across different global or historic contexts. Other critiques shine light on additional concerns about the validity of StereoSet's data in terms of aspects ranging from spelling and grammar to the inaccuracy of claims that the stereotypes represented in the benchmark actually reflect harmful biases rather than innocuous biases or contextual ambiguities \cite{Govil_2025}. 

Therefore, while our findings demonstrate bias spillovers \textit{within the StereoSet framework}, future work is essential to validate these trade-offs in real-world applications and across more culturally-aware benchmarks. First, we will expand to newer social bias benchmarks such as BBQ \cite{parrish2022bbq}, which highlights attested biases against people belonging to protected classes along nine social dimensions. 
Furthermore, complementary to bias benchmarks like CrowS-Pairs and StereoSet, RealToxicityPrompts \cite{gehman2020realtoxicityprompts} targets generative toxicity, providing prompts and scoring methods to quantify how frequently language models produce toxic continuations in realistic settings: it will be well worth exploring whether other forms of alignment, e.g., harm mitigation, could lead to unintended exacerbation of other harm dimensions.

\section{Conclusions}
Our study systematically investigated the cross-category effects of targeted bias mitigation techniques in LLMs, presenting a framework for thorough bias analysis as well as compelling evidence for our posited "No Free Lunch" principle in debiasing. By applying four distinct post-hoc bias mitigation methods (Logit Steering, Activation Patching, BiasEdit, and Prompt Debiasing) across ten transformer-based LLMs and evaluating their impact on four dimensions of bias (gender, profession, religion, and race) using the StereoSet benchmark, we revealed a pattern of systemic trade-offs. 

Our results show that \textbf{targeted debiasing frequently leads to collateral damage}, significantly worsening overall model utility in the majority of experimental runs. Additionally, we find that bias mitigation techniques seeking to address only one dimension of bias often result in cross-category spillover. In many cases, out findings demonstrate that \textbf{mitigating bias along a singular dimension can exacerbate it along another}. While there are many existing debiasing techniques that vary in terms of efficacy, the methods we explored consistently have the potential to \textbf{unintentionally make models more biased and even less coherent overall.} In addition, while we found that larger models are more robust and equipped to resist the occasionally catastrophic effects of debiasing on general language modeling abilities, bias mitigation techniques varied significantly across model types. This suggests that \textbf{many characteristics of biases and how they manifest in model architectures are unexplored}. 

Our findings also lay the foundation for future approaches to fairer, more robust methods for model debiasing. Firstly, existing methods are designed to mitigate single-axis biases that occur within the context of singular simplistic sentences. Models generally display lower understanding of longer contexts and intersentence data overall. Yet, natural language and societal biases are not constrained by sentence length and complexity. Future methods will need to be equipped to handle both longer, intersentence debiasing data and more complex, multi-dimensional, and even intersectional biases in order to accurately represent language and society. 

Another future route for investigation is how biases manifest differently in models with varying architectures. As models' complexity increases, this represents an important step in the fields of both bias research and mechanistic interpretability. 

Finally, future benchmarks and tools for bias detection and analysis must prioritize compositional balance and clear contextual validity: it is essential for tools claiming to detect certain harmful biases to ground their data in real-world harms experienced by varying communities rather than examples that may misrepresent the very biases they purport to address. This balance will not only bolster the framework we have presented but also make significant progress toward fair LLMs.  

Ultimately, our work highlights the "No Free Lunch" principle: \textbf{every debiasing intervention comes with trade-offs}. Someday, these insights may guide the development of LLMs and, more broadly, AI systems that balance trade-offs thoughtfully to become fairer, more reliable, and more reflective of the communities they serve.

\bibliography{references} 


\end{document}